\documentclass[sigconf]{acmart}

\AtBeginDocument{%
\providecommand\BibTeX{{%
\normalfont B\kern-0.5em{\scshape i\kern-0.25em b}\kern-0.8em\TeX}}}
    
\usepackage{xcolor}
\usepackage{colortbl}
\usepackage{multirow}
\usepackage{enumitem}
\usepackage{dblfloatfix}
\usepackage{subcaption}

\usepackage{booktabs}

\usepackage{pgfplots}
\pgfplotsset{compat=1.12}
\usepackage{filecontents}
\usepgfplotslibrary{statistics}
\usetikzlibrary{pgfplots.statistics, pgfplots.colorbrewer} 
\usepackage{pgfplotstable}

\newcommand\red[1]{{\color{red}#1}} 
\newcommand\blue[1]{{\color{blue}#1}}
\newcommand\new[1]{{\color{black}#1}}

\begin{document}

\title{Adversarial attacks to image classification systems using evolutionary algorithms}

\author{Sergio Nesmachnow}
\affiliation{
    \institution{Universidad de la República}
    \country{Uruguay}
}
\email{sergion@fing.edu.uy}

\author{Jamal Toutouh}
\affiliation{
\institution{ITIS, Universidad de Málaga}
\country{Spain}
}
\email{jamal@lcc.uma.es}

\begin{abstract}
Image classification currently faces significant security challenges due to adversarial attacks, which consist of intentional alterations designed to deceive classification models based on artificial intelligence. 
This article explores an approach to generate adversarial attacks against image classifiers using a combination of evolutionary algorithms and generative adversarial networks. The proposed approach explores the latent space of a generative adversarial network with an evolutionary algorithm to find vectors representing adversarial attacks. The approach was evaluated in two case studies corresponding to the classification of handwritten digits and object images. The results showed success rates of up to 35\% for handwritten digits, and up to 75\% for object images, improving over other search methods and reported results in related works. The applied method proved to be effective in handling data \new{diversity on the target datasets}, even in problem instances that presented additional challenges due to the complexity and richness of information.

\end{abstract}

%
%



\maketitle

\section{Introduction}

In the field of machine learning, image classification has emerged as a cornerstone application with significant impact across various technology fields and industries, including security, healthcare, and personalized marketing~\cite{Szeliski2011}. Image recognition and classification systems play a vital role in applications designed to locate objects, identify individuals, and detect features in images. However, challenges such as improving precision and enhancing robustness have become key in advancing towards fully functional, reliable, and independent image recognition systems~\cite{Singh2020}. Among these challenges, the vulnerability of image classification methods to adversarial attacks has gained increasing attention~\cite{Vakhshiteh2021}.

An adversarial attack involves deliberately altering input data to mislead a machine learning model into making incorrect predictions~\cite{Vakhshiteh2021}. The induced errors pose serious risks to the security and privacy of critical systems, e.g., surveillance and public safety systems~\cite{thys2019fooling}.

This article presents an approach for generating adversarial attacks against image classifiers using a combination of evolutionary algorithms (EAs) and Generative Adversarial Networks (GANs). 
GANs learn data distributions to generate synthetic data that closely resembles real samples. GANs have been applied in many scientific and commercial fields, especially for generating synthetic images and videos~\cite{Gui2023}. 
The methodology is based on searching the latent space of GANs using EAs to find suitable vectors for generating images representing adversarial attacks against specific image classifiers. 
This approach advances over state-of-the-art models in literature~\cite{Clare2023} by proposing and analyzing two new fitness functions explicitly designed to optimize adversarial attacks, which balance classifier confusion and misclassification rates.
Two case studies are addressed: generating attacks on classifiers for handwritten digits and object images. The proposed approach was designed to create effective \new{(i.e., able to deceive the classifier)}, diverse, and high-quality adversarial examples to assess the robustness of classifiers in various problem variants. Furthermore, this method is flexible and can be applied to other datasets and classifiers, as long as a trained generative model is available to produce data samples. 

The obtained results showed the effectiveness of the proposed approach in generating adversarial attacks against image classifiers, achieving competitive success rates across different problems, and significantly improving the baseline method. For handwritten digits, successful adversarial examples were generated for all classes, with the highest success rate reaching 35\%. For object images, the approach performed better, achieving a peak success rate of 75\%.

The article is organized as follows. Next section describes the addressed problem and methodology. The description of the proposed EA for generating adversarial attacks to classifiers is presented in Section~\ref{Sec:EA}. Section~\ref{Sec:Attacks_MNIST_CIFAR} describes the application of the evolutionary search to generate adversarial attacks against image classifiers for handwritten digits and object images. The experimental analysis and results are reported in Section~\ref{Sec:Experiments}. Finally, Section~\ref{Sec:Conclusions} presents the conclusions and formulates the main lines for future work.

\section{Adversarial attacks via latent space search of GANs}
\label{Sec:Problem}

This section presents the considered problem and the methodology for generating adversarial attacks via latent space search of GANs.

\subsection{Adversarial attacks}

Adversarial attacks are subtle and intentional alterations to the input data of a machine learning model, with the purpose of deceiving the system and obtaining incorrect responses to jeopardize the reliability and accuracy of the model~\citep*{explaining_adversarial_examples}. 
In image classification, adversarial attacks 
can manifest as images with perturbations that 
lead image classifiers to incorrectly identify an object or identity, {or to produce ambiguous predictions where the two most likely classes are too similar, under a given threshold $\delta$}. Such scenarios can have serious implications for the security and privacy of critical systems, for example, in surveillance and public safety systems.
%
Continuous research in generating new adversarial attacks is essential for image classification systems to stay updated and resilient against emerging attack techniques, ensuring their security, accuracy, and reliability in changing environments~\cite{Chakraborty2021}.

GANs are artificial neural networks (ANNs) specialized in learning the distributions, features, and labels of an input dataset of real data, with the main goal of generating new synthetic data samples that follow a distribution approximated to the one of real data~\cite{Foster2019}. 
GANs apply adversarial training between two ANNs: a generator,
which is 
trained to create new synthetic samples taking latent space vectors as input, and a discriminator, which learns to distinguish between real and synthetic samples while providing feedback to improve the generator. 
The ultimate goal of the generator is to approximate the real data distribution and produce synthetic data that is virtually indistinguishable from real data to deceive the discriminator.

The latent space of GANs is a continuous multidimensional space
sampled from a random distribution
(e.g., Gaussian or uniform). 
Traditional 
gradient-based search 
methods often struggle to navigate it effectively due to its high dimensionality and lack of clear structure for determining useful directions.
Thus, EAs and other metaheuristics have been applied to guide the search and find useful vectors for the problem at hand~\cite{Machin2022}. EAs take advantage of their high versatility to deal with different GAN architectures and latent space features and their robustness to deal with 
changing and noisy 
optimization functions. 
The black-box optimization approach applied by EAs allows using different 
surrogate 
functions to guide the search, without relying on gradient-based operators~\cite{Volz2018}.

The overall strategy 
\new{considers a set of latent search vectors in the population}.
The evolutionary cycle apply the traditional selection and variation operators. The hybridization with the generative approach is performed on the fitness evaluation of candidate vectors: the conditional GAN is applied to generate images that are then evaluated using different classifiers. The fitness function is defined according to different metrics that allow identifying successful attacks. This way, by applying evolutionary computation, a robust global search strategy is defined for the exploration of the latent space of the considered conditional GAN to generate synthetic images that successfully attack the evaluated classifiers.

To develop an effective method for generating adversarial attacks, certain requirements needed to be established to ensure the efficiency and robustness of the process. 
\new{The first requirement (R1) is to ensure that the generated adversarial attack images have a high visual quality. The second requirement (R2) is that the generated attacks should be correctly classified by the human eye. Finally, the third requirement (R3) is generating diverse attacks and ensuring that the adversarial examples have varied characteristics}.

The proposed 
EA 
for generating adversarial attacks is evaluated on two 
image classification datasets: MNIST (Modified National Institute of Standards and Technology) database~\cite{Li2012} and CIFAR (Canadian Institute For Advanced Research) 10 dataset~\cite{Darlow2018}.

\subsection{Related work}

Several recent articles have addressed the generation of adversarial attacks via the exploration of the latent space of generative models. Trajectory-based methods have been applied for attack generation in natural language processing models~\cite{Liu2022,Li2023}, images~\cite{Fan2024,Xiao2018,Meiner2023}, and other representations~\cite{Chang2022}. Approaches have applied deterministic search, e.g. via gradient descent, or random perturbations. Deterministic search is computationally efficient, but the search is highly dependable on the initial candidate solution, the method may get stuck in local optima and is not applicable to non-differentiable or discontinuous search functions. Other approaches have applied surrogate models and direct manipulation of synthetic data~\cite{Papernot2017}. 

Population-based methods have shown improved accuracy, but many black-box approaches for adversarial attacks have relied on specific constraints or assumptions~\cite{Brunner2019,Chang2022,Andriushchenko2020} or applied heuristic algorithms~\cite{Li2023}. 
Specific approaches have been proposed to bias the search to improve the efficiency of black-box methods for generating adversarial attacks~\cite{Brunner2019}. 
Alzantot et al.~\cite{Alzantot2019} applied a genetic search for black-box generation of attacks, but only exploring perturbations instead of searching the full latent space. 
\new{These 
methods operate in the pixel space and have shown high accuracy in black-box adversarial attacks, achieving competitive results across various configurations. However, despite their performance, they exhibit important limitations: they lack semantic guidance, are often limited to untargeted attacks, and do not leverage latent representations or apply evolutionary principles in a structured search space.}

The use of multiple fitness functions for attack generation has also been explored~\cite{Wu2021}. \new{The paper reported over 60\% success on CIFAR-10 using a multi-objective GA, though the absence of a generative model limited the realism and generality of the attacks.}

Closed to our research, Clare and Correia~\cite{Clare2023} generated adversarial attacks via latent space exploration using a fitness function that evaluated the proximity to the distribution of real data. A second stage was needed to evaluate if the generated samples were effective attacks or not. \new{Their method achieved 25--30\% success on Fashion-MNIST and CIFAR-10, but required post-processing, which limited integration and efficiency.}

\new{In this line of work, our article contributes a compact and flexible framework for adversarial attack generation that combines semantic latent exploration with efficient evolutionary search, using simple variation operators and adaptable fitness designs. Our 
EA
achieved competitive results, without requiring gradients, surrogate models, or post-filtering stages. }

\section{Evolutionary algorithm for adversarial attacks to classifiers}
\label{Sec:EA}

This section describes the approach applying EAs for generating adversarial attacks to classifiers.


\textit{Solution Encoding}.
Each solution is encoded as a vector of floating-point numbers, representing a specific point in the latent space of the applied GAN. The dimensionality of these vectors corresponds to the input dimension required by the generator.

\textit{Initialization}.
The population is initialized using a stochastic procedure, where each value in the solution vector is sampled from a normal distribution $\mathcal{N}(0, 1)$. Preliminary experiments confirmed that this simple stochastic initialization provides sufficient diversity in the initial population, enabling effective exploration of the latent space. No prior knowledge of specific features or latent space directions is required to begin the evolutionary search.

\textit{Selection}.
The tournament selection operator was applied. The parameters of the tournament selection were configured to three participants and one winner. Initial experiments showed that these settings provided a correct selection pressure to guide the evolutionary search for adversarial attacks effectively. 

\textit{Recombination}.
A two-point crossover operator was applied, which showed a better recombination pattern in preliminary experiments compared to a one-point crossover and arithmetic crossover.

\textit{Mutation}.
A Gaussian mutation operator was applied, with a mean $\mu = 0$ and standard deviation $\sigma = 1$. This operator effectively balanced maintaining and introducing diversity in the population while minimizing disruption to the search process.

\textit{Replacement}.
A $\mu$+$\lambda$ replacement strategy was applied
to maintain 
diversity in the population, providing a proper balance between exploration and exploitation, and rapidly finding accurate solutions.

\section{Adversarial attacks to classifiers of image datasets}
\label{Sec:Attacks_MNIST_CIFAR}

This section details the application of the proposed 
EA 
to generate adversarial attacks, evaluated on two standard image classification datasets: MNIST and CIFAR-10. The datasets, generative models, classifiers, and fitness functions are described below.

\subsection{Datasets}
\label{Sec:Datasets}

The MNIST~\cite{Li2012} dataset comprises 70,000 grayscale images of handwritten digits. It consists of 60,000 training and 10,000 test images 28$\times$28 pixels in size each.
This dataset is widely used for benchmarking machine learning methods in classification tasks. 

The CIFAR-10 dataset~\cite{Darlow2018} consists of 60,000 color images, categorized into 10 classes (airplane, automobile, bird, cat, deer, dog, frog, horse, ship, and truck). CIFAR-10 consists of a total of 60,000 examples, with 50,000 images for training and 10,000 for evaluation, where each image has a size of 32$\times$32 pixels. This dataset is more complex than MNIST because the images are in color and exhibit more significant intra-class variability. 

\subsection{Generative models and classifiers}
\label{Sec:Models_Classifiers}

The approach for generating adversarial attacks through latent space search 
applies
a generative model to create attacks and a classifier to recognize 
if
the generated image is an attack or not.

Conditional GANs (CGANs) were chosen for their ability to generate diverse, class-specific samples—an essential feature for targeted attacks. 
Among publicly available and open-source CGANs, models that produced high-quality visual samples were selected. For each dataset, the chosen CGAN generates samples that maximize their classification accuracy of state-of-the-art classifiers. Thus, a series of preliminary experiments were carried out to choose the generators. This approach ensured that the generated adversarial examples were both visually convincing and effective for testing the robustness of classifiers. 

\new{The generator used for MNIST is based on Conditional Deep Convolutional GAN 
by Mirza and Osindero~\cite{Mirza2014}. This model uses a 100-dimensional normally distributed latent space to generate 28$\times$28 grayscale images of digits. 
In turn, the generative model for CIFAR-10 relied on Energy-based Conditional GAN~\cite{Chen2021},
which
has a latent space of dimension 80 and produces 32$\times$32 color images.} 

\new{The classifiers used to evaluate the attacks and to guide the search are the publicly available ones that provided the highest classification accuracy on the training dataset in preliminary experiments. For MNIST, classifier $c_{C}$ is based on a multi-layer perceptron that achieved over 99\% accuracy~\cite{Bose2019}. For CIFAR-10, classifier $c_{C}$ is based on a ResNet56 architecture that achieved 94.4\% accuracy~\cite{Wang2023}, sufficient for evaluating the generated adversarial attacks.}



\subsection{Fitness functions}
\label{sec:funciones-de-fitness}

Two different fitness functions were studied for the evaluation of candidate solutions. These fitness functions require minor adjustments to accommodate dataset-specific characteristics. The fitness functions consider the following elements:

\begin{itemize}
    \item A latent space of dimension $d$, \( \mathbb{Z} \) $\subseteq$ \( \mathbb{R}^{d} \) 
    \item \( \mathbb{I} \subseteq \mathbb{R}^{s \times s} \) the image space (where $s \times s$ is the image size)
    \item \( \mathbb{K} = \{k_0, k_1, \ldots, k_l\} \) the set of $l$ class labels
    \item $k$ $\in$ $\mathbb{K}$ the target label to attack
    \item \( g : \mathbb{Z}, \mathbb{K} \rightarrow \mathbb{I} \) the generative model
    \item \( \mathbb{P} \subseteq [0,1]^{l} \) the probabilities assigned to each label in $\mathbb{K}$
    \item {A classifier \( c : \mathbb{I} \rightarrow \mathbb{P} \), where \( c_{k} : \mathbb{I} \rightarrow [0,1] \) is the probability assigned by classifier $c$ to label $k$}
\end{itemize}


Fitness function $f_1$
addresses adversarial attack generation by maximizing the confusion (or minimizing the confidence) in the predictions. It evaluates the extent to which the classifier avoids confidently assigning high probabilities to any label for a generated sample, including the target label  $k$. A higher value of $f_1$ 
means
greater confusion in the classifier, indicating that the generated sample successfully reduces the confidence across all possible labels.


 \begin{equation}
    f_1(z) = 1 - \max_{p \in {\mathbb{P}}} c(g(z,k))
    \label{Eq:fitness1}
\end{equation}


Fitness function $f_2$
aims to create a scenario where the classifier is uncertain about its prediction, reducing the likelihood of correctly classifying the generated adversarial attack. It minimizes the difference between the probabilities assigned to the target predicted label $p$ and the second most likely label, forcing the classifier to struggle between them. Besides, it minimizes the probability of the target label $k$. A higher $f_2$ value reflects increased confusion and reduced confidence in the predictions of the classifier.


\begin{equation}
\begin{split}
    f_2(z) & = 1 - \left| \max_{p \in {\mathbb{P}}} c(g(z,k)) - \max_{q \in {\mathbb{P}} \setminus \{p\}} c(g(z,k)) \right| \\
    & + 1 - c_{k}(g(z,k))
    \label{Eq:fitness2}
\end{split}
\end{equation}

These functions enabled the generation of adversarial images that significantly \new{challenged the robustness of the classifier}. The specific values for $d$, $l$, and $c$ are 100, 10, and $c_{M}$ for MNIST, respectively, and 80, 10, and $c_{C}$ for CIFAR-10.

\begin{table*}[!h]
\setlength{\belowcaptionskip}{0pt}
\setlength{\abovecaptionskip}{3pt}
\renewcommand{\arraystretch}{0.85}
\centering
\footnotesize 
\caption{Number of adversarial attacks generated for MNIST classifier $c_{M}$, grouped by target digit and fitness function} 
\label{Table:attacks_clas-1}
\resizebox{\textwidth}{!}{  
\begin{tabular}{ll*{10}{>{\centering\arraybackslash}p{0.92cm}} >{\centering\arraybackslash}p{1.2cm}} 
\toprule
\multirow{2}{*}{\centering fitness} & & \multicolumn{10}{c}{\textit{target digit}} & \multirow{2}{*}{\centering\textbf{total}} \\
\cmidrule{3-12}
 & & 0 & 1 & 2 & 3 & 4 & 5 & 6 & 7 & 8 & 9 & \\
\midrule
$f_1$ & & 38\,098 & 19\,624 & 37\,742 & 58\,512 & 55\,661 & 57\,647 & 10\,409 & 34\,411 & 43\,550 & 3\,918 & \textbf{359\,572} \\
\midrule
$f_2$ & & 46\,386 & 31\,555 & 51\,813 & 108\,399 & 63\,309 & 96\,490 & 9\,622 & 50\,141 & 63\,860 & 1\,698 & \textbf{523\,273} \\
\bottomrule
\end{tabular}
}
\end{table*}

\begin{table*}[!h]
\setlength{\tabcolsep}{3pt}
\setlength{\belowcaptionskip}{0pt}
\setlength{\abovecaptionskip}{3pt}
\renewcommand{\arraystretch}{0.85}
\centering
\caption{Number of correctly classified instances in which the two highest probabilities provided by classifier $c_{M}$ are within a distance less than $\delta$, grouped by target digit of attack and fitness function} 
\label{table:mnist-attacks-under-threshold}
\begin{tabular}{c}
    \begin{tabular}{c*{11}{@{\hskip 0.4cm}>{\centering\arraybackslash}p{0.8cm}}>{\centering\arraybackslash}p{1.6cm}@{\hskip 0.1cm}}
    \toprule
    \multicolumn{1}{c}{\multirow{2}{*}{\textit{fitness}}} & \multicolumn{1}{c}{\multirow{2}{*}{\textit{$\delta$}}} & \multicolumn{10}{c}{\textit{target digit}} & \multicolumn{1}{c}{\multirow{2}{*}{\textbf{total}}} \\
    \cmidrule(lr){3-12}
    & & 0 & 1 & 2 & 3 & 4 & 5 & 6 & 7 & 8 & 9 \\
    \midrule    
    \multirow{5}{*}{\textit{$f_1$}} & $<0.5$ & 27\,194 & 39\,181 & 25\,482 & 25\,596 & 24\,166 & 25\,959 & 22\,692 & 31\,227 & 32\,935 & 9\,811 & \textbf{264\,243} \\
    & $<0.4$ & 20\,927 & 31\,333 & 19\,773 & 20\,553 & 19\,968 & 20\,722 & 16\,959 & 24\,418 & 26\,161 & 7\,022 & \textbf{207\,836} \\
    & $<0.3$ & 15\,288 & 23\,721 & 14\,616 & 15\,610 & 15\,793 & 15\,668 & 11\,946 & 18\,179 & 19\,831 & 4\,736 & \textbf{155\,388}\\
    & $<0.2$ & 10\,075 & 16\,142 & 9\,884 & 10\,790 & 11\,382 & 10\,772 & 7\,582 & 12\,246 & 13\,568 & 2\,924 & \textbf{105\,365}\\
    & $<0.1$ & 5\,278 & 8\,543 & 5\,509 & 5\,907 & 6\,450 & 5\,882 & 3\,751 & 6\,676 & 7\,322 & 1\,478 & \textbf{56\,796}\\
    \bottomrule
    \end{tabular} 
\end{tabular}

\begin{tabular}{c}
    \begin{tabular}{c*{11}{@{\hskip 0.4cm}>{\centering\arraybackslash}p{0.8cm}}>{\centering\arraybackslash}p{1.6cm}@{\hskip 0.1cm}}
    & \multicolumn{1}{c}{\multirow{2}{*}{\textit{$\delta$}}} & \multicolumn{10}{c}{\textit{target digit}} & \multicolumn{1}{c}{\multirow{2}{*}{\textbf{total}}} \\
    \cmidrule(lr){3-12}
    & & 0 & 1 & 2 & 3 & 4 & 5 & 6 & 7 & 8 & 9 \\
    \midrule
    \multirow{5}{*}{\textit{$f_2$}} & $<0.5$ & 22\,593 & 21\,311 & 20\,236 & 9\,617 & 15\,687 & 11\,252 & 25\,616 & 23\,886 & 20\,771 & 11\,679 & \textbf{182\,648} \\
    & $<0.4$ & 17\,258 & 15\,401 & 15\,383 & 7\,593 & 12\,271 & 9\,010 & 19\,264 & 18\,459 & 16\,434 & 8\,296 & \textbf{139\,369}\\
    & $<0.3$ & 12\,581 & 10\,519 & 11\,120 & 5\,707 & 9\,113 & 6\,759 & 13\,860 & 13\,340 & 12\,366 & 5\,396 & \textbf{100\,761}\\
    & $<0.2$ & 8\,267 & 6\,396 & 7\,226 & 3\,959 & 6\,103 & 4\,649 & 8\,898 & 8\,698 & 8\,312 & 3\,031 & \textbf{65\,539}\\
    & $<0.1$ & 4\,320 & 2\,978 & 3\,739 & 2\,119 & 3\,196 & 2\,494 & 4\,444 & 4\,371 & 4\,349 & 1\,255 & \textbf{33\,265}\\
    \bottomrule
    \end{tabular} 
\end{tabular}
\end{table*}

\subsection{Implementation}
The proposed 
EA 
was implemented using Python 3.9 and the PyGAD open-source library for evolutionary and machine learning algorithms (\url{https://pygad.readthedocs.io/}). PyGAD provides 
support for building and training ANNs using 
EAs.

PyGAD allows customizing each step of the proposed evolutionary approach for generating adversarial attacks, enabling a simple experimentation and facilitating the incorporation of specific components such as the classifiers and the conditional GANs used for both case studies. The implementation of the proposed method for generating adversarial attacks is available in the public repository 
\url{gitlab.fing.edu.uy/sergion/ataques-adversarios-con-algoritmos-evolutivos-y-redes-generativas-antagonicas}.
The experimental evaluation was performed on the high-performance computing infrastructure of the 
National Supercomputing Center (Cluster-UY) in Uruguay~\cite{Nesmachnow2019}.

\section{Experimental evaluation}
\label{Sec:Experiments}

This section describes the empirical analysis of the proposed EAs for generating adversarial attacks. 
All images generated during the search are stored to be evaluated later.

\subsection{Parameters setting}
\label{Sec:parameter-setting}
For both addressed datasets, the studied parameters included the population size ($\#P$), the number of generations ($\#g$) used as stopping criterion, the recombination probability ($p_R$), and the mutation probability ($p_M$).
Other parameters, including the tournament size, the values of $\mu = 2$, and $\lambda = 1$, were set in preliminary experiments. 

Candidate values for studied parameters were 
$\#P$ in \{50,100,200\}, $\#g$ in \{200,300,400\}, $p_R$ in \{0.60,0.75,0.90\}, and $p_M$ in \{10$^{-3}$,10$^{-2}$,10$^{-1}$\}. 
Each parameter configuration was evaluated 
on 30 independent executions 
for the evaluated fitness functions and problem instances.

The Friedman rank statistical test was applied to analyze the distributions.
The best
results were computed using 
$\#P = 50$, $\#g = 400 $, $p_R = 0.75$ and $p_M = 10^{-1}$ for MNIST.
In contrast, for CIFAR-10, results of the Friedman rank statistical test confirmed that the 
best 
results were computed using the configuration $\#P = 100$, $\#g = 400 $, $p_R = 0.9$ and $p_M = 10^{-1}$, i.e., a greater population size and a higher value of $p_R$ were needed. The higher complexity and details of the images in the CIFAR-10 dataset required a deeper exploration of the latent space than for the MNIST dataset.

\subsection{Fitness evolution}
\label{Sec:evolution}

\begin{figure}[!ht]
\setlength{\abovecaptionskip}{3pt}
\setlength{\belowcaptionskip}{-15pt}
    \centering
\includegraphics[width=0.75\linewidth]{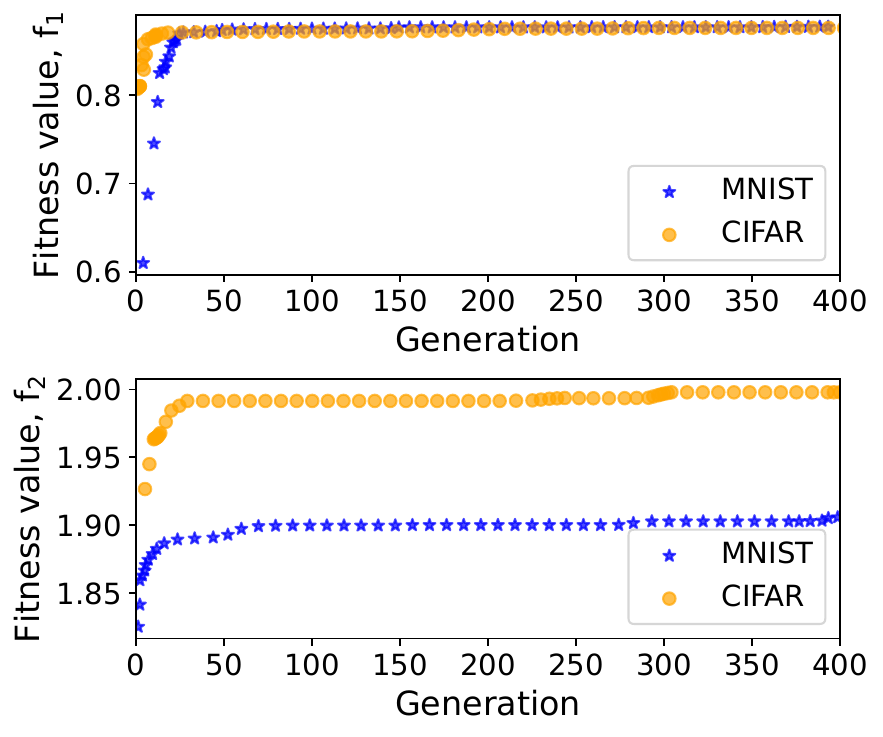}
\caption{Mean fitness evolution for MNIST and CIFAR-10}
\label{fig:fitness_evolution}
\end{figure}

Figure~\ref{fig:fitness_evolution} shows the evolution of the mean fitness value 
for $f_1$ (top) and $f_2$ (bottom) 
for both datasets.

\new{$f_1$ increased rapidly for both datasets, rising 
within the first 50 generations} and plateauing afterward. 
%
\new{$f_1$ showed similar trends on both datasets because $f_1$ does not exploit dataset complexity or inter-class variability (CIFAR-10 is more complex and has more inter-class variability than MNIST). The goal of $f_1$ is to maximize overall classifier confusion without targeting class similarities.} 
\new{In contrast, $f_2$ exhibited distinct trends for the two datasets, emphasizing its more targeted optimization approach.} 
For MNIST, $f_2$ increased steadily 
over 150 generations, reflecting the simpler nature of this dataset and the slower process of reducing classifier confidence. 
For CIFAR-10, $f_2$ rapidly increased 
in 50 generations, 
stabilizing afterward. The faster convergence shows a higher complexity of CIFAR-10, which provides more opportunities for $f_2$ to create ambiguity between the two most probable classes. Unlike $f_1$, $f_2$ leverages dataset-specific features to generate more precise adversarial examples, leading to distinct performance differences across datasets. 
\new{The smoother MNIST curves and noisier CIFAR-10 patterns suggest a more rugged fitness landscape in the latter, likely due to greater visual variability.}

\subsection{Attacks to handwritten digits classifiers}
\label{Sec:MNIST-resultst}

The evaluation 
performed
30 independent executions of the proposed EA for each digit,
using the studied fitness functions. 

Table~\ref{Table:attacks_clas-1} reports the number of attacks generated for each 
digit using 
$f_1$ and $f_2$ against classifier $c_{M}$. \new{The total number of attacks for each class are resorted in bold font.}
%
%
Out of 
{1\,500\,000 generated images},
24\% were attacks 
using 
$f_1$ and 35\% using 
$f_2$.
%
%
Digits 3, 4, and 5 were the most susceptible to adversarial attacks, with over 50\,000 attacks each. In contrast, digits 6 and 9 were more challenging 
(fewer than 15\,000 attacks).
{The disparity may arise from the visual similarity of digits 6 and 9 to other classes, which may complicate the generation of effective perturbations.}  Overall, 
$f_2$ produced more attacks across most digits than $f_1$, but digits 6 and 9.
\new{The significantly large number of attacks found demonstrate the usefulness of the proposed approach, as this number is to be maximized.}

\new{All generated images had a high visual quality (R1) and were correctly classifiable by the human eye (R2).}
Regarding 
correctly classified images (\new{R3, the assigned label matched the ground truth})
\new{an attack was considered successful if the two highest probabilities were within a 
distance $\delta$, indicating confusion between classes}. Table~\ref{table:mnist-attacks-under-threshold} presents the number of generated images meeting this condition, grouped by 
$\delta$ and fitness function.


More than 50\,000 examples were generated where the difference between the two highest probabilities was less than 0.1 using 
$f_1$, and more than 30\,000 using 
$f_2$. This finding suggests that 
the EA 
also produced samples that are correctly classified but still confuse the classifier.
{Although $f_2$ was specifically designed to produce confusion between the two most probable classes, its higher overall success rate in generating misclassified attacks lowers the whole correctly produced images. Consequently, $f_2$ produced fewer attacks of correctly classified examples compared to $f_1$.} 


Digit 3 was
analyzed in detail as 
it
exhibited the highest number of attacks. 
Table~\ref{Table:attacks_to_3} displays
the number of attacks on digit 3, grouped by fitness function, 
{probability thresholds 
set to classify a prediction as an attack, and the label provided by classifier $c_{M}$.

\begin{table}[!h]
\setlength{\tabcolsep}{2pt}
\setlength{\belowcaptionskip}{0pt}
\setlength{\abovecaptionskip}{3pt}
\renewcommand{\arraystretch}{0.85}
\caption{Number of attacks to digit 3}
\label{Table:attacks_to_3}
\centering
    \begin{tabular}{c*{10}{>{\centering\arraybackslash}p{0.8cm}}}
    \toprule
    \multicolumn{1}{c}{\multirow{2}{*}{\textit{fitness}}} & \multicolumn{1}{c}{\multirow{2}{*}{\textit{p}}} & \multicolumn{6}{c}{\textit{class}}
    & \multicolumn{2}{>{\hskip 0.2cm}c}{\multirow{2}{*}{\textbf{Total}}}\\
    \cmidrule(lr){3-8}
    & & 0 & 2 & 5 & 7 & 8 & 9 \\
    \midrule
    \multirow{5}{*}{\textit{$f_1$}} & $>0$ & 688 & 5\,235 & 12\,985 & 923 & 6\,960 & 31\,721 & \multicolumn{2}{>{\hskip 0.2cm}c}{\textbf{58\,512}}\\
    & $>0.5$ & 440 & 4\,034 & 8\,987 & 548 & 4\,344 & 25\,985 & \multicolumn{2}{>{\hskip 0.2cm}c}{\textbf{44\,338}}\\
    & $>0.6$ & 308 & 2\,976 & 6\,133 & 356 & 2\,897 & 20\,198 & \multicolumn{2}{>{\hskip 0.2cm}c}{\textbf{32\,868}}\\
    & $>0.7$ & 178 & 2\,092 & 3\,907 & 203 & 1\,816 & 15\,272 & \multicolumn{2}{>{\hskip 0.2cm}c}{\textbf{23\,468}}\\
    & $>0.8$ & 108 & 1\,310 & 2\,212 & 110 & 1\,034 & 10\,848 & \multicolumn{2}{>{\hskip 0.2cm}c}{\textbf{15\,622}}\\
    & $>0.9$ & 45 & 635 & 949 & 43 & 433 & 6\,314 & \multicolumn{2}{>{\hskip 0.2cm}c}{\textbf{8\,419}} \\
    \bottomrule
    & \multicolumn{1}{c}{\multirow{2}{*}{\textit{p}}} & \multicolumn{4}{c}{\textit{class}} & \multicolumn{2}{>{\hskip 0.2cm}c}{\multirow{2}{*}{\textbf{Total}}} \\
    \cmidrule(lr){3-6}
    & & 2 & 5 & 8 & 9 \\
    \midrule
    \multirow{5}{*}{\textit{$f_2$}} & $>0$ & 3\,039 & 36\,660 & 4\,916 & 63\,784 & \multicolumn{2}{>{\hskip 0.2cm}c}{\textbf{108\,399}} \\
    & $>0.5$ & 2\,762 & 34\,178 & 4\,577 & 61\,025 & \multicolumn{2}{>{\hskip 0.2cm}c}{\textbf{102\,542}} \\
    & $>0.6$ & 2\,035 & 25\,618 & 3\,621 & 49\,768 & \multicolumn{2}{>{\hskip 0.2cm}c}{\textbf{81\,042}} \\
    & $>0.7$ & 1\,320 & 17\,870 & 2\,878 & 39\,210 & \multicolumn{2}{>{\hskip 0.2cm}c}{\textbf{61\,278}} \\
    & $>0.8$ & 739 & 10\,755 & 2\,185 & 28\,406 & \multicolumn{2}{>{\hskip 0.2cm}c}{\textbf{42\,085}} \\
    & $>0.9$ & 241 & 4\,450 & 1\,517 & 16\,768 & \multicolumn{2}{>{\hskip 0.2cm}c}{\textbf{22\,976}} \\
    \bottomrule \\[-12pt]
\end{tabular}
\end{table}

\begin{table}[!ht]
\centering
\setlength{\tabcolsep}{3pt}
\setlength{\abovecaptionskip}{3pt}
\renewcommand{\arraystretch}{0.85}
\caption{Sample attacks to classifier $c_{M}$ (digit 3)}
\label{Table:sample_attacks_to_3}
\footnotesize 
\begin{tabular}{lrrrrrrrrrr}
\toprule
 {image} & \multicolumn{10}{l}{\includegraphics[width=193px]{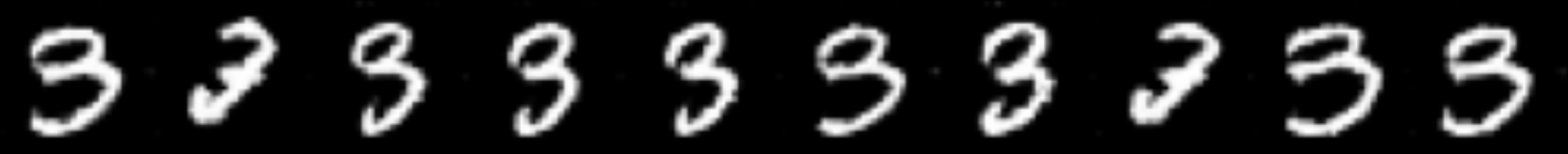}} \\
 \midrule
 class & 9 & 9 & 9 & 9 & 9 & 9 & 9 & 9 & 9 & 9 \\
 \midrule
 probability & $0.99$ & \hspace{0mm}$0.99$ & \hspace{0mm}$0.99$ & \hspace{1mm}$0.99$ & \hspace{1mm}$0.99$ & \hspace{0mm}$0.99$ & \hspace{0mm}$0.98$ & \hspace{1mm}$0.98$ & \hspace{1mm}$0.98$ & \hspace{1mm}$0.99$ \\
 \bottomrule \\[-12pt]
\end{tabular}
\end{table}

\begin{table}[!h]
\centering
\setlength{\tabcolsep}{3pt}
\setlength{\abovecaptionskip}{3pt}
\renewcommand{\arraystretch}{0.85}
\caption{Sample attacks to classifier $c_{M}$ (digit 9)}
\label{Table:sample_attacks_to_9}
\footnotesize 
\begin{tabular}{lrrrrrrrrrr}
\toprule
 {image} & \multicolumn{10}{l}{\includegraphics[width=193px]{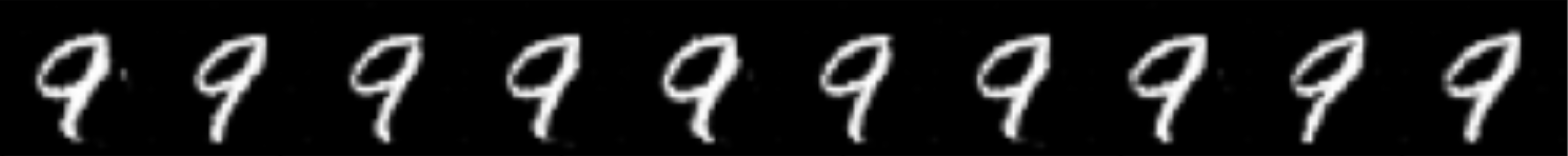}} \\
 \midrule
 class & 7 & 7 & 7 & 7 & 7 & 7 & 7 & 7 & 7 & 7 \\
 \midrule
 probability & $0.96$ & \hspace{0mm}$0.97$ & \hspace{0mm}$0.97$ & \hspace{0mm}$0.97$ & \hspace{1mm}$0.97$ & \hspace{1mm}$0.97$ & \hspace{0mm}$0.98$ & \hspace{0mm}$0.98$ & \hspace{1mm}$0.98$ & \hspace{-1mm}$0.99$ \\
 \bottomrule \\[-15pt]
\end{tabular}
\end{table}

\begin{table*}[!b]
\setlength{\tabcolsep}{3.75pt} 
\setlength{\abovecaptionskip}{3pt}
\setlength{\belowcaptionskip}{0pt}
\renewcommand{\arraystretch}{0.85}
\centering
\caption{Number of adversarial attacks generated for CIFAR-10 classifier $c_{C}$, grouped by target digit and fitness function} 
\label{Table:attacks_cifar_clas-1}
\begin{tabular}{ll*{10}{>{\centering\arraybackslash}p{0.92cm}} >{\centering\arraybackslash}p{1.2cm}} 
\toprule
\multirow{2}{*} & & \multicolumn{10}{c}{\textit{target object}} & \multirow{2}{*}{\centering\textbf{Total}} \\
\cmidrule{3-12}
& & \multicolumn{1}{c}{airplane} & \multicolumn{1}{c}{car} & \multicolumn{1}{c}{bird} & \multicolumn{1}{c}{cat} & \multicolumn{1}{c}{deer} & \multicolumn{1}{c}{dog} & \multicolumn{1}{c}{frog} & \multicolumn{1}{c}{horse} & \multicolumn{1}{c}{ship} & \multicolumn{1}{c}{truck} \\
\midrule
$f_1$ & & 305\,578 & 247\,784 & 216\,923 & 219\,213 & 195\,167 & 268\,045 & 285\,000 & 176\,786 & 251\,710 & 155\,062 & \textbf{2\,321\,268} \\
\midrule
$f_2$ & & 358\,333 & 314\,822 & 274\,273 & 288\,058 & 273\,521 & 343\,726 & 292\,980 & 257\,261 & 344\,986 & 242\,424 & \textbf{2\,990\,348} \\
\bottomrule
\end{tabular}
\end{table*}
\begin{table*}[!b]
\setlength{\tabcolsep}{2.5pt}
\setlength{\abovecaptionskip}{3pt}
\renewcommand{\arraystretch}{0.85}
\caption{Number of correctly classified instances in the CIFAR-10 dataset in which the two highest probabilities are within a distance less than $\delta$, grouped by target object and fitness function.}
\label{table:cifar-real-instances}
\centering
\begin{tabular}{c}
    \begin{tabular}{c*{11}{@{\hskip 0.4cm}>{\centering\arraybackslash}p{0.8cm}}>{\centering\arraybackslash}p{1.6cm}@{\hskip 0.1cm}}
    \toprule
    \multicolumn{1}{c}{\multirow{2}{*}{\textit{fitness}}} & \multicolumn{1}{c}{\multirow{2}{*}{$\delta$}} & \multicolumn{10}{c}{\textit{target object}} & \multicolumn{1}{c}{\multirow{2}{*}{\textbf{Total}}} \\
    \cmidrule(lr){3-12}
    & & airplane & car & bird & cat & deer & dog & frog & horse & ship & truck \\
    \midrule
    \multirow{5}{*}{\textit{$f_1$}} & $<0.5$ & 40\,349 & 43\,749 & 100\,378 & 53\,517 & 48\,118 & 38\,989 & 49\,264 & 39\,930 & 54\,978 & 63\,024 & \textbf{532\,296} \\
    & $<0.4$ & 36\,139 & 37\,978 & 97\,170 & 45\,548 & 41\,121 & 33\,714 & 45\,936 & 33\,347 & 47\,744 & 52\,232 & \textbf{470\,929}\\
    & $<0.3$ & 31\,548 & 31\,893 & 93\,575 & 37\,746 & 34\,092 & 28\,352 & 42\,625 & 26\,599 & 40\,061 & 41\,498 & \textbf{407\,989}\\
    & $<0.2$ & 26\,003 & 25\,113 & 88\,941 & 29\,362 & 26\,295 & 22\,605 & 38\,943 & 19\,535 & 31\,256 & 30\,033 & \textbf{338\,086}\\
    & $<0.1$ & 18\,557 & 16\,270 & 78\,505 & 19\,370 & 16\,898 & 15\,563 & 33\,518 & 11\,436 & 19\,956 & 17\,561 & \textbf{247\,634}\\
    \bottomrule
    \end{tabular}
\end{tabular}
\begin{tabular}{c}
    \begin{tabular}{c*{11}{@{\hskip 0.4cm}>{\centering\arraybackslash}p{0.8cm}}>{\centering\arraybackslash}p{1.6cm}@{\hskip 0.1cm}}
    \multicolumn{1}{c}{\multirow{2}{*}{\textit{fitness}}} & \multicolumn{1}{c}{\multirow{2}{*}{$\delta$}} & \multicolumn{10}{c}{\textit{target object}} & \multicolumn{1}{c}{\multirow{2}{*}{\textbf{Total}}} \\
    \cmidrule(lr){3-12}
    & & airplane & car & bird & cat & deer & dog & frog & horse & ship & truck \\
    \midrule
    \multirow{5}{*}{\textit{$f_2$}} & $<0.5$ & 10\,665 & 15\,233 & 20\,096 & 21\,385 & 18\,522 & 10\,811 & 16\,064 & 18\,874 & 13\,061 & 31\,853 & \textbf{176\,564} \\
    & $<0.4$ & 8\,673 & 12\,331 & 16\,232 & 17\,000 & 14\,823 & 8\,692 & 12\,771 & 14\,920 & 10\,517 & 25\,351 & \textbf{141\,310} \\
    & $<0.3$ & 6\,672 & 9\,459 & 12\,614 & 12\,761 & 11\,223 & 6\,539 & 9\,668 & 11\,131 & 8\,093 & 19\,160 & \textbf{107\,320} \\
    & $<0.2$ & 4\,591 & 6\,512 & 8\,746 & 8\,574 & 7\,696 & 4\,446 & 6\,557 & 7\,499 & 5\,512 & 12\,956 & \textbf{73\,089} \\
    & $<0.1$ & 2\,412 & 3\,468 & 4\,714 & 4\,356 & 3\,994 & 2\,326 & 3\,500 & 3\,823 & 2\,846 & 6\,714 & \textbf{38\,153}\\
    \bottomrule
    \end{tabular}
\end{tabular}
\end{table*}

\begin{table*}[!b]
\setlength{\tabcolsep}{2.5pt}
\setlength{\abovecaptionskip}{3pt}
\renewcommand{\arraystretch}{0.85}
\centering
\caption{Number of attacks to airplane object in the CIFAR-10 dataset}
\label{table:ataques-al-object-avion}
\begin{tabular}{c}
    \begin{tabular}{c*{11}{@{\hskip 0.4cm}>{\centering\arraybackslash}p{0.8cm}}>{\centering\arraybackslash}p{1.6cm}}
    \toprule
    \multicolumn{1}{c}{\multirow{2}{*}{\textit{fitness}}} & \multicolumn{1}{c}{\multirow{2}{*}{\textit{p}}} & \multicolumn{9}{c}{\textit{class}} & \multicolumn{2}{c}{\multirow{2}{*}{\textbf{Total}}} \\
    \cmidrule(lr){3-11}
    & & car & bird & cat & deer & dog & frog & horse & ship & truck \\
    \midrule
    \multirow{5}{*}{\textit{$f_1$}} & $>0$ & 4\,835 & 97\,149 & 93\,824 & 25\,691 & 4\,933 & 33\,536 & 13\,242 & 24\,494 & 7\,874 & \multicolumn{2}{c}{\textbf{305\,578}} \\
    & $>0.5$ & 885 & 24\,973 & 26\,033 & 13\,683 & 696 & 9\,716 & 2\,825 & 3\,815 & 669 & \multicolumn{2}{c}{\textbf{83\,295}} \\
    & $>0.6$ & 590 & 20\,738 & 19\,376 & 10\,971 & 519 & 7\,779 & 1\,762 & 2\,908 & 465 & \multicolumn{2}{c}{\textbf{65\,108}} \\
    & $>0.7$ & 406 & 17\,315 & 14\,521 & 8\,637 & 376 & 6\,104 & 1\,079 & 2\,158 & 334 & \multicolumn{2}{c}{\textbf{50\,930}} \\
    & $>0.8$ & 268 & 14\,050 & 10\,538 & 6\,502 & 259 & 4\,663 & 635 & 1\,473 & 232 & \multicolumn{2}{c}{\textbf{38\,620}} \\
    & $>0.9$ & 135 & 10\,333 & 6\,863 & 4\,183 & 163 & 3\,125 & 276 & 867 & 132 & \multicolumn{2}{c}{\textbf{26\,077}} \\
    \bottomrule
    \end{tabular}
\end{tabular}
\begin{tabular}{c}
    \begin{tabular}{c*{11}{@{\hskip 0.4cm}>{\centering\arraybackslash}p{0.8cm}}>{\centering\arraybackslash}p{1.6cm}}
    \multicolumn{1}{c}{\multirow{2}{*}{\textit{fitness}}} & \multicolumn{1}{c}{\multirow{2}{*}{\textit{p}}} & \multicolumn{9}{c}{\textit{class}} & \multicolumn{2}{c}{\multirow{2}{*}{\textbf{Total}}} \\
    \cmidrule(lr){3-11}
    & & car & bird & cat & deer & dog & frog & horse & ship & truck \\
    \midrule
    \multirow{5}{*}{\textit{$f_2$}} & $>0$ & 397 & 88\,872 & 112\,568 & 52\,009 & 4\,037 & 44\,359 & 32\,935 & 20\,904 & 2\,252 & \multicolumn{2}{c}{\textbf{358\,333}} \\
    & $>0.5$ & 250 & 66\,615 & 84\,345 & 40\,044 & 2\,240 & 31\,573 & 23\,107 & 16\,021 & 1\,549 & \multicolumn{2}{c}{\textbf{265744}} \\
    & $>0.6$ & 186 & 57\,024 & 70\,743 & 33\,546 & 1\,816 & 27\,122 & 17\,501 & 13\,467 & 1\,159 & \multicolumn{2}{c}{\textbf{222\,564}} \\
    & $>0.7$ & 131 & 48\,404 & 58\,312 & 27\,643 & 1\,443 & 23\,219 & 12\,746 & 11\,083 & 851 & \multicolumn{2}{c}{\textbf{183\,832}} \\
    & $>0.8$ & 80 & 39\,881 & 46\,036 & 21\,974 & 1\,128 & 19\,061 & 8\,677 & 8\,734 & 579 & \multicolumn{2}{c}{\textbf{146\,150}} \\
    & $>0.9$ & 39 & 29\,776 & 32\,193 & 15\,572 & 732 & 14\,041 & 4\,789 & 6\,014 & 335 & \multicolumn{2}{c}{\textbf{103\,491}} \\
    \bottomrule
    \end{tabular}
\end{tabular}
\end{table*}

A high vulnerability 
was evident for digit 3, as a wide variety of successful attacks were generated, even with high probability thresholds. 
The classifier confused 
digit 3 with digits 0, 2, 5, 7, 8, and 9. 
The attacks where the classifier assigned a probability greater than 0.9 to the incorrect digit were 8,419 when using fitness function $f_1$ and 22,976 when using fitness function 
$f_2$. In more than 70\% of these attacks, the classifier confused digit 3 with digit 9. Table~\ref{Table:sample_attacks_to_3} presents ten sample attacks generated for digit 3. Table~\ref{Table:sample_attacks_to_9} presents ten sample attacks generated for digit 9, the most challenging digit.

\subsection{Attacks to classifiers of common objects}
\label{Sec:CIFAR-resultst}

Thirty independent executions of the proposed EA were performed
for each object in 
CIFAR-10
using the studied fitness functions.

\begin{table*}[!t]
\centering
\setlength{\tabcolsep}{4.25pt}
\setlength{\abovecaptionskip}{3pt}
\renewcommand{\arraystretch}{0.85}
\caption{Sample attacks to classifier $c_{C}$ with the class airplane}
\label{table:ejemplos-ataques-airplane}
\begin{tabular}{llccccccccccc}
\toprule
 \multicolumn{2}{l}{image} & \multicolumn{11}{l}{\includegraphics[width=310px]{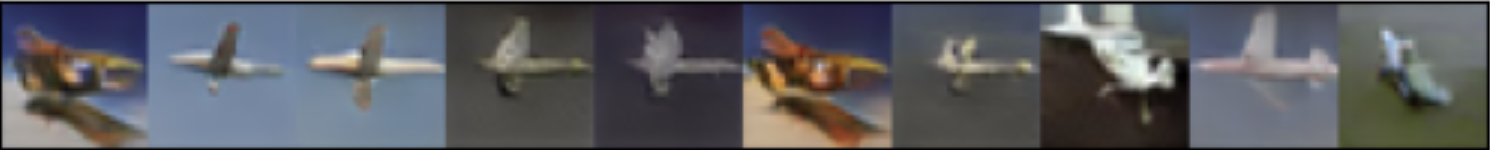}} \\
 \midrule
 class & & \makebox[0.8cm]{car} & \makebox[0.85cm]{bird} & \makebox[0.85cm]{bird} & \makebox[0.85cm]{cat} & \makebox[0.8cm]{deer} & \makebox[0.8cm]{dog} & \makebox[0.8cm]{frog} & \makebox[0.75cm]{horse} & \makebox[0.8cm]{ship} & \makebox[0.8cm]{truck} \\
  \midrule
 probability & & $0.99$ & $0.99$ & $0.99$ & $0.99$ & $0.99$ & $0.99$ & $0.99$ & $0.99$ & $0.99$ & $0.99$ & \\
\bottomrule \\[-12pt]
\end{tabular}
\end{table*}

\begin{table*}[!t]
\centering
\setlength{\tabcolsep}{4.25pt}
\setlength{\abovecaptionskip}{3pt}
\renewcommand{\arraystretch}{0.85}
\caption{Sample attacks to classifier $c_{C}$ with the class truck}
\label{table:ejemplos-ataques-truck}
\begin{tabular}{llccccccccccc}
\toprule
 \multicolumn{2}{l}{image} & \multicolumn{11}{l}{\includegraphics[width=310px]{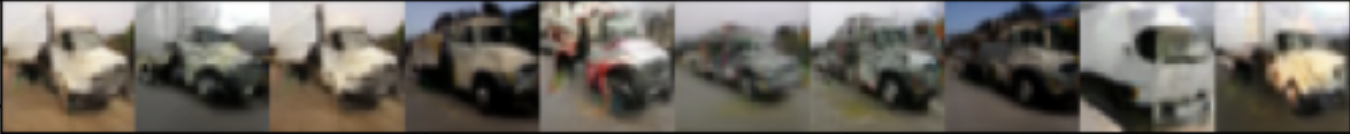}} \\
 \midrule
 class & & \makebox[0.8cm]{plane} & \makebox[0.8cm]{car} & \makebox[0.85cm]{bird} & \makebox[0.85cm]{bird} & \makebox[0.85cm]{cat} & \makebox[0.8cm]{deer} & \makebox[0.8cm]{dog} & \makebox[0.8cm]{frog} & \makebox[0.75cm]{horse} & \makebox[0.8cm]{ship} \\
  \midrule
 probability & & $0.99$ & $0.99$ & $0.93$ & $0.99$ & $0.99$ & $0.99$ & $0.97$ & $0.88$ & $0.99$ & $0.99$ & \\
\bottomrule \\[-12pt]
\end{tabular}
\end{table*}

Table~\ref{Table:attacks_cifar_clas-1} reports the number of attacks generated for each target object using 
$f_1$ and $f_2$ against classifier $c_{C}$. 
%
\new{All generated images had a high visual quality (R1) and were correctly classifiable by the human eye (R2).}
Out of all generated images,
58\% were attacks against classifier $c_{C}$ using 
$f_1$ and 75\% using 
$f_2$,
significantly higher than for MNIST.
More than 150\,000 attacks were generated for each object. 
A greater number of attacks were obtained using 
$f_2$ compared to $f_1$ for each object. 
{This
result
confirms that using $f_2$, which explicitly focuses on generating confusion between the two most probable classes,
is better
to generate
adversarial attacks. 
By promoting ambiguity and lowering the likelihood of correct predictions, $f_2$ effectively exploits classifier vulnerabilities.}
%
Airplane was the most susceptible class to adversarial attacks, with over than 650\,000 attacks. 
Car, horse, and deer
were more challenging,
with lower than 500\,000 attacks.

Table~\ref{table:cifar-real-instances} reports the number of \new{correctly classified instances where the two highest probabilities had a difference less than a certain threshold $\delta$}. 
%
More than 240\,000 examples were generated with a difference between the two highest probabilities was less than 0.1 using 
$f_1$, and more than 30\,000 using 
$f_2$ (similar to the results 
for MNIST, 
where $f_1$ produced more attacks of this type than $f_2$). This finding suggests that this method for generating adversarial attacks can also produce examples that are correctly classified but still confuse the classifier{, mainly using fitness function $f_1$}.

Attacks to the object airplane 
were analyzed in detail, as 
it
represents the 
extreme
in the observed distribution. 
More than 300\,000 attacks were generated on the object airplane, 
suggesting it is 
easy to attack.
Table~\ref{table:ataques-al-object-avion} displays the number of attacks on the airplane object, organized according to the fitness function used, different probability thresholds set to classify a prediction as an attack, and the label provided by the classifier $c_{C}$. 

The classifier confused the images of airplanes 
with all objects in the CIFAR-10 dataset. For the airplane, 26\,077 attacks were obtained where the highest assigned probability was greater than 0.9 when using fitness function $f_1$, and 103\,491 attacks when using fitness function $f_2$. In both cases, the classifier confused the generated airplane with a bird or a cat in over 60\% of the attacks. 
Table~\ref{table:ejemplos-ataques-airplane} presents ten sample attacks generated for the airplane class. Table~\ref{table:ejemplos-ataques-truck} presents ten sample attacks generated for the truck class.


The results obtained in the evaluation of CIFAR-10 were highly positive, since multiple attacks were successfully generated for all objects in the dataset. In turn, the findings suggest that image classifiers for objects are more vulnerable than classifiers for handwritten digits. This difference could be attributed to the fact that object image classifiers require higher resolution in the processed images to distinguish details accurately.

\subsection{Comparison with a multistart local search}

The proposed EA was compared with a multistart iterated local search (MILS) method based on Alzantot et al.~\cite{Alzantot2019}, but extended to explore the whole latent space and only perturbations. 

MiLS
applies the 
Gaussian mutation operator for exploring in the neighborhood of the current solution, using 
$\mu = 0$ and $\sigma = 1$.
A predefined effort stopping criterion is applied, performing the same number of function evaluations as the proposed EA. To avoid getting stuck in a local optima, a reinitialization is applied if no improvement is found in 1000 evaluations~\cite{Iturriaga2014}.

Tables~\ref{Table:attacks_comp_EA_MILS_MNIST} and~\ref{Table:attacks_comp_EA_MILS_CIFAR} report the number of attacks generated by the compared search methods for MNIST and CIFAR-10 datasets using function $f_2$\new{, which allowed to compute the larger number of attacks in the experiments performed.}
Results show that the proposed EA generated significantly more attacks than MILS. Improvements ranged from
8.01\% (digit 1) to 61.43\% (digit 6) for MNIST and from 22.49\% (dog) to 35.58\% (car) for CIFAR-10. 

\begin{table}[!h]
\setlength{\belowcaptionskip}{0pt}
\setlength{\abovecaptionskip}{3pt}
\renewcommand{\arraystretch}{0.85}
\setlength{\tabcolsep}{6pt}
\centering
\caption{EA vs.~MILS: Number of attacks to classifier $c_{M}$ (MNIST) using $f_2$}
\label{Table:attacks_comp_EA_MILS_MNIST}
\begin{tabular}{lrr}
\toprule
\textit{Target digit} & \textit{Attacks (EA/MILS)} & \textit{EA over MILS} \\
\midrule
0 & 46\,386/35\,003 & 24.54\% \\
1 & 31\,555/29\,029 & 8.01\% \\
2 & 51\,813/37\,481 & 27.66\% \\
3 & 108\,399/88\,217 & 18.62\% \\
4 & 63\,309/50\,315 & 20.52\% \\
5 & 96\,490/80\,047 & 17.04\% \\
6 & 9\,622/3\,711 & 61.43\% \\
7 & 50\,141/37\,206 & 25.80\% \\
8 & 63\,860/48\,288 & 24.38\% \\
9 & 1\,698/720 & 57.60\% \\
\midrule
\textbf{Total} & \textbf{523\,273/410\,017} & \textbf{21.64\%} \\
\bottomrule \\[-12pt]
\end{tabular}
\end{table}

\begin{table}[!h]
\setlength{\belowcaptionskip}{0pt}
\setlength{\abovecaptionskip}{3pt}
\renewcommand{\arraystretch}{0.85}
\setlength{\tabcolsep}{6pt}
\centering
\caption{EA vs.~MILS: Number of attacks to classifier $c_{C}$ (CIFAR-10) using $f_2$}
\label{Table:attacks_comp_EA_MILS_CIFAR}
\begin{tabular}{lrr}
\toprule
\textit{Target object} & \textit{Attacks (EA/MILS)} & \textit{EA over MILS} \\
\midrule
airplane & 358\,333/260\,121 & 27.41\% \\
car & 314\,822/202\,803 & 35.58\% \\
bird & 274\,273/183\,390 & 33.14\% \\
cat & 288\,058/204\,501 & 29.01\% \\
deer & 273\,521/209\,691 & 23.34\% \\
dog & 343\,726/266\,413 & 22.49\% \\
frog & 292\,980/227\,048 & 22.50\% \\
horse & 257\,261/198\,503 & 22.84\% \\
ship & 344\,986/261\,582 & 24.18\% \\
truck & 242\,424/179\,413 & 25.99\% \\
\midrule
\textbf{Total} & \textbf{2\,990\,348/2\,193\,465} & \textbf{26.65\%} \\
\bottomrule \\[-12pt]
\end{tabular}
\end{table}

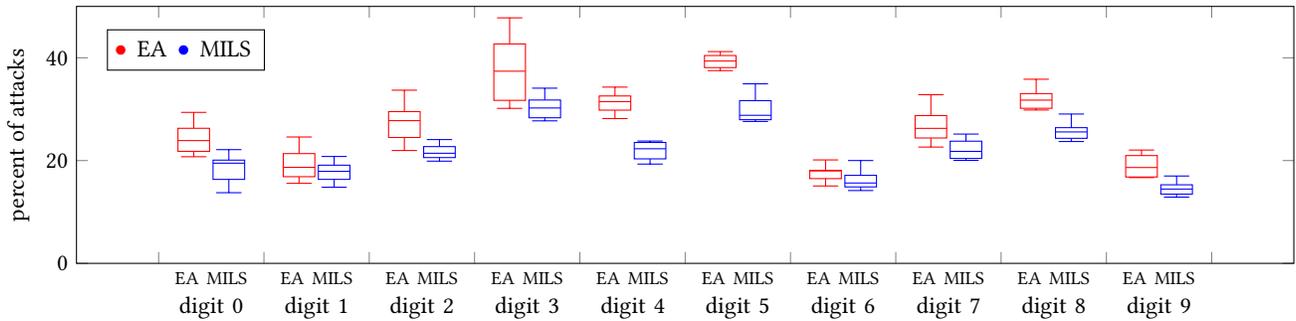
\begin{figure*}[!h]
\setlength{\abovecaptionskip}{3pt}
\setlength{\belowcaptionskip}{-9pt}
    \centering
\begin{tikzpicture}
\pgfplotsset{
  /pgfplots/custom legend/.style={
    legend image code/.code={
      \draw [only marks,draw=white,mark=*]
      plot coordinates {(0.3cm,0cm)};
    },
  },    
}
\begin{axis}[
boxplot/draw direction=y,
ylabel={percent of attacks},
height=5cm,
ymin=0,ymax=50,
cycle list={{red},{blue}},
boxplot={
        %
        draw position={1/3 + floor(\plotnumofactualtype/2) + 1/3*mod(\plotnumofactualtype,2)}, 
        %
        box extend=0.3
},
x=1.4cm,
xtick={0,1,2,...,20},
x tick label as interval,
xticklabels={%
        {{\footnotesize EA MILS}}\\digit 0,%
        {{\footnotesize EA MILS}}\\digit 1,%
        {{\footnotesize EA MILS}}\\digit 2,%
        {{\footnotesize EA MILS}}\\digit 3,%
        {{\footnotesize EA MILS}}\\digit 4,%
        {{\footnotesize EA MILS}}\\digit 5,%
        {{\footnotesize EA MILS}}\\digit 6,%
        {{\footnotesize EA MILS}}\\digit 7,%
        {{\footnotesize EA MILS}}\\digit 8,%
        {{\footnotesize EA MILS}}\\digit 9,%
},
        x tick label style={
                text width=2.5cm,
                align=center
        },
  custom legend,
  legend style={
    legend columns=2,
    column sep=0.25em,
    at={(0.025,0.75)}, anchor=south west
  },
]

\addlegendentry{EA}
\addlegendentry{MILS}

\addplot 
table[row sep=\\,y index=0] {
data\\
27.64\\
29.38\\
20.75\\
24.93\\
22.85\\
};

\addplot
table[row sep=\\,y index=0] {
data\\
13.75\\
22.13\\
20.06\\
18.93\\
20.12\\
};

\addplot 
table[row sep=\\,y index=0] {
data\\
19.21\\
24.59\\
15.57\\
23.51\\
18.16\\
};

\addplot
table[row sep=\\,y index=0] {
data\\
20.80\\
14.82\\
17.87\\
20.26\\
17.95\\
};

\addplot 
table[row sep=\\,y index=0] {
data\\
27.03\\
33.70\\
30.55\\
28.51\\
21.96\\
};

\addplot
table[row sep=\\,y index=0] {
data\\
21.28\\
21.57\\
24.09\\
23.87\\
19.89\\
};

\addplot 
table[row sep=\\,y index=0] {
data\\
47.78\\
43.88\\
30.16\\
33.28\\
41.51\\
};

\addplot
table[row sep=\\,y index=0] {
data\\
32.03\\
31.54\\
34.09\\
27.73\\
28.93\\
};

\addplot 
table[row sep=\\,y index=0] {
data\\
31.45\\
28.18\\
34.30\\
33.66\\
31.49\\
};

\addplot
table[row sep=\\,y index=0] {
data\\
28.93\\
19.32\\
21.35\\
23.80\\
23.25\\
};

\addplot 
table[row sep=\\,y index=0] {
data\\
40.07\\
41.21\\
38.67\\
40.81\\
37.51\\
};

\addplot
table[row sep=\\,y index=0] {
data\\
29.32\\
27.61\\
34.94\\
28.30\\
34.01\\
};

\addplot 
[boxplot={draw position = 19/3}, color = red]
table[row sep=\\,y index=0] {
data\\
18.19\\
17.92\\
15.06\\
20.13\\
18.03\\
};

\addplot
table[row sep=\\,y index=0] {
data\\
20.01\\
14.17\\
15.70\\
18.56\\
15.58\\
};

\addplot 
table[row sep=\\,y index=0] {
data\\
26.37\\
32.82\\
31.15\\
26.17\\
22.63\\
};

\addplot
table[row sep=\\,y index=0] {
data\\
20.80\\
22.78\\
25.16\\
24.77\\
20.04\\
};

\addplot 
[boxplot={draw position = 25/3}, color = red]
table[row sep=\\,y index=0] {
data\\
30.51\\
29.84\\
35.83\\
33.02\\
33.04\\
};

\addplot
table[row sep=\\,y index=0] {
data\\
29.06\\
26.62\\
24.96\\
23.71\\
26.19\\
};

\addplot 
table[row sep=\\,y index=0] {
data\\
16.69\\
22.05\\
21.57\\
16.92\\
20.38\\
};

\addplot
table[row sep=\\,y index=0] {
data\\
14.85\\
12.89\\
15.73\\
14.07\\
17.00\\
};
\end{axis}
\end{tikzpicture}
\caption{Comparison of attack success rates: EA vs.~MILS on MNIST}
\label{Fig:boxplots_EA_MILS_MNIST}
\end{figure*}

\begin{figure*}[!h]
\setlength{\abovecaptionskip}{3pt}
\setlength{\belowcaptionskip}{-9pt}
\centering
\begin{tikzpicture}
\pgfplotsset{
  /pgfplots/custom legend/.style={
    legend image code/.code={
      \draw [only marks,draw=white,mark=*]
      plot coordinates {(0.3cm,0cm)};
    },
  },    
}
\begin{axis}[
boxplot/draw direction=y,
ylabel={percent of attacks},
height=5cm,
ymin=0,ymax=80,
cycle list={{red},{blue}},
boxplot={
        %
        draw position={1/3 + floor(\plotnumofactualtype/2) + 1/3*mod(\plotnumofactualtype,2)}, 
        %
        box extend=0.3
},
x=1.4cm,
xtick={0,1,2,...,20},
x tick label as interval,
xticklabels={%
        {{\footnotesize EA MILS}}\\airplane,%
        {{\footnotesize EA MILS}}\\car,%
        {{\footnotesize EA MILS}}\\bird,%
        {{\footnotesize EA MILS}}\\cat,%
        {{\footnotesize EA MILS}}\\deer,%
        {{\footnotesize EA MILS}}\\dog,%
        {{\footnotesize EA MILS}}\\frog,%
        {{\footnotesize EA MILS}}\\horse,%
        {{\footnotesize EA MILS}}\\ship,%
        {{\footnotesize EA MILS}}\\truck,%
},
        x tick label style={
                text width=2.5cm,
                align=center
        },
          custom legend,
  legend style={
    legend columns=2,
    column sep=0.25em,
    at={(0.025,0.15)}, anchor=south west
  },
]

\addlegendentry{EA}
\addlegendentry{MILS}

\addplot 
table[row sep=\\,y index=0] {
data\\
64.17\\
69.80\\
60.75\\
74.83\\
72.25\\
};

\addplot
table[row sep=\\,y index=0] {
data\\
53.75\\
52.13\\
60.06\\
58.93\\
60.12\\
};

\addplot 
table[row sep=\\,y index=0] {
data\\
59.21\\
64.59\\
65.57\\
63.51\\
58.16\\
};

\addplot
table[row sep=\\,y index=0] {
data\\
40.80\\
48.82\\
52.87\\
40.26\\
37.95\\
};

\addplot 
table[row sep=\\,y index=0] {
data\\
47.03\\
53.70\\
60.55\\
58.51\\
51.96\\
};

\addplot
table[row sep=\\,y index=0] {
data\\
36.28\\
31.57\\
34.09\\
26.87\\
29.89\\
};

\addplot 
table[row sep=\\,y index=0] {
data\\
27.78\\
33.88\\
30.16\\
32.80\\
31.51\\
};

\addplot
table[row sep=\\,y index=0] {
data\\
12.03\\
11.54\\
14.09\\
17.73\\
18.93\\
};

\addplot 
table[row sep=\\,y index=0] {
data\\
37.45\\
30.18\\
28.30\\
35.66\\
31.49\\
};

\addplot
table[row sep=\\,y index=0] {
data\\
18.93\\
19.32\\
21.35\\
20.80\\
23.25\\
};

\addplot 
table[row sep=\\,y index=0] {
data\\
40.07\\
46.21\\
38.67\\
43.81\\
39.51\\
};

\addplot
table[row sep=\\,y index=0] {
data\\
26.32\\
24.61\\
24.94\\
28.30\\
30.01\\
};

\addplot 
[boxplot={draw position = 19/3}, color = red]
table[row sep=\\,y index=0] {
data\\
38.19\\
37.92\\
35.06\\
40.13\\
38.03\\
};

\addplot
table[row sep=\\,y index=0] {
data\\
25.01\\
24.17\\
25.70\\
22.56\\
20.58\\
};

\addplot 
table[row sep=\\,y index=0] {
data\\
26.37\\
34.82\\
37.15\\
26.17\\
29.63\\
};

\addplot
table[row sep=\\,y index=0] {
data\\
20.80\\
22.78\\
15.16\\
24.77\\
20.04\\
};

\addplot 
[boxplot={draw position = 25/3}, color = red]
table[row sep=\\,y index=0] {
data\\
50.51\\
49.84\\
45.83\\
43.02\\
43.04\\
};

\addplot
table[row sep=\\,y index=0] {
data\\
39.06\\
36.62\\
34.96\\
33.71\\
36.19\\
};

\addplot 
table[row sep=\\,y index=0] {
data\\
36.69\\
32.05\\
31.57\\
36.92\\
31.38\\
};

\addplot
table[row sep=\\,y index=0] {
data\\
24.85\\
22.89\\
25.73\\
24.07\\
27.00\\
};
\end{axis}
\end{tikzpicture}
\caption{Comparison of attack success rates: EA vs.~MILS on CIFAR-10}
\label{Fig:boxplots_EA_MILS_CIFAR}
\end{figure*}
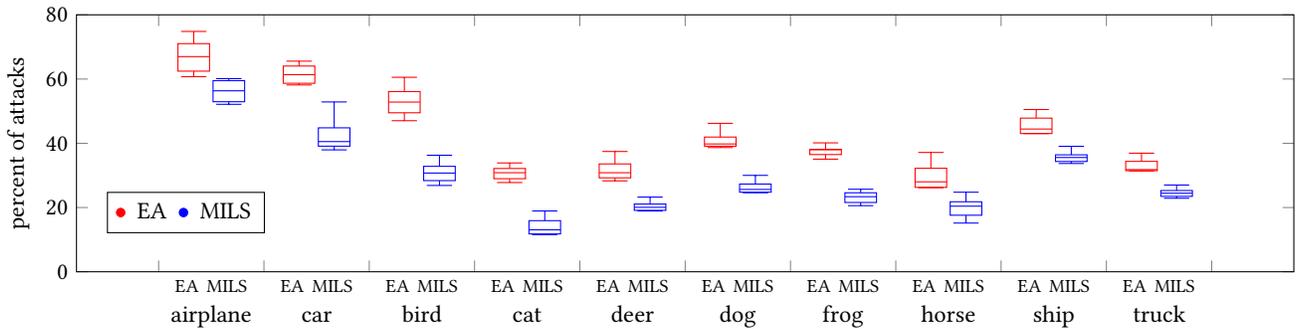

Figures~\ref{Fig:boxplots_EA_MILS_MNIST} and ~\ref{Fig:boxplots_EA_MILS_CIFAR} present the boxplot comparison between EA and MILS for each class. 
The proposed EA improved over MILS for all classes in both case studies. Boxplots indicate that MNIST is easier to solve than CIFAR-10, as MILS computed similar results than the proposed EA for digits 1 and 6, where the improvements of EA were smaller than the inter-quartile range
of the results distributions. 

Other digits, such as 3 and 5, were easier to attack using EA, and significant improvements over MILS are reported. Regarding CIFAR, the specific features of images on the dataset made it harder for a simple local search method to find attacks. The improvements of the proposed EA over MILS were statistically significant for all classes. The higher improvements were computed for bird, whereas deer and truck 
had
the smaller improvements of EA over MILS.

\new{Figure~\ref{Fig:fitness_evolution_EA_vs_MILS} presents representative graphics of the average evolution of fitness function $f_2$ for EA and MILS on MNIST}. The proposed EA showed rapid convergence, reaching a fitness value of 0.87 within 50 generations, 
whereas
MILS had a slower progression, eventually plateauing at a lower value of approximately 0.79. 
The fitness evolution patterns
highlight the ability of the EA to explore the latent space and maximize classifier confusion, achieving faster convergence speed and better final fitness value. }

\new{Regarding the comparison with results from the related literature, the proposed EA was competitive with results for similar problems. Wu et al.~\cite{Wu2021} reported success rates above 60\% on CIFAR-10 using a multi-objective GA in the perturbation space, while Clare and Correia~\cite{Clare2023} achieved 25--30\% using a two-stage latent space approach. Our method reached up to 75\% success on CIFAR-10 and 35\% on MNIST, using a simpler, fully integrated evolutionary framework with no gradient, surrogate model, or post-processing requirements.}

\begin{figure}[!ht]
\setlength{\abovecaptionskip}{3pt}
\setlength{\belowcaptionskip}{-12pt}
    \centering
\includegraphics[width=0.85\linewidth]{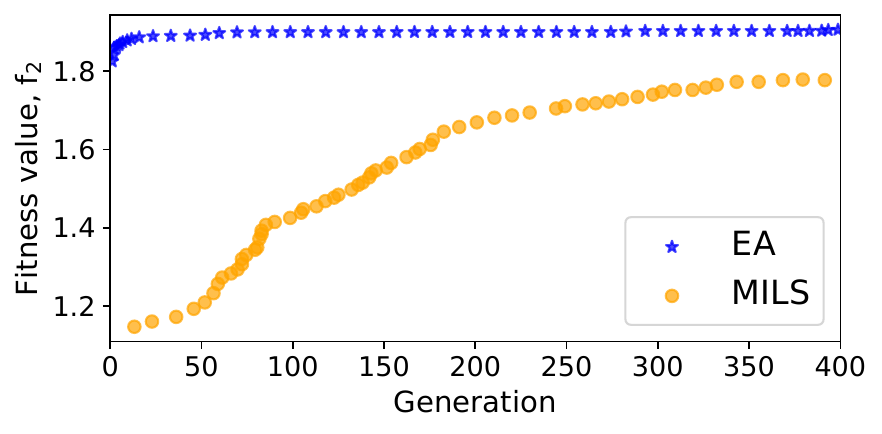}
\caption{Fitness evolution: EA vs.~MILS}
\label{Fig:fitness_evolution_EA_vs_MILS}
\end{figure}

\section{Conclusions and Future Work}
\label{Sec:Conclusions}

This article presented an approach to generating adversarial attacks on image classifiers by leveraging the latent space of GANs through EAs, addressing a critical challenge in improving the robustness of recognition and classification systems. The evolutionary search employed a black-box approach guided by two novel fitness functions designed to balance classifier confusion and misclassification.

Experimental results on MNIST and CIFAR-10 datasets demonstrated the effectiveness of the proposed approach, achieving success rates 
up to 75\%,
\new{a remarkable success rate compared to related works,}
and significantly outperforming a MILS method. 
The EA showed
a more consistent and fast evolution pattern for both studied fitness functions. 
The fitness function 
considering confusion between the two most probable labels allowed to generate more attacks than the one considering confidence across all possible labels.
The findings revealed that object classifiers, such as those trained on CIFAR-10, are more vulnerable to adversarial attacks than simpler classifiers like those used for handwritten digits.

This article contributes to the field by demonstrating the potential of integrating EAs with GAN-based latent space exploration, offering a flexible framework for testing classifier resilience. \new{EAs leverage their adaptability to work with various GAN architectures and latent space characteristics, along with their resilience in handling partial information thanks to the black-box optimization they applied. These features define a proper exploration pattern that allows improving other traditional methods.} The adaptability of the approach allows its application to other datasets and classifier architectures, providing a valuable tool for enhancing the robustness of machine learning systems. \new{The versatility of EAs makes the approach applicable to other generation and classification problems, including text, audio, and natural language.}

The main lines for future work are related to extending the experimental validation of the proposed approach and designing more powerful variation operators. \new{We also propose applying the methodology to the human faces recognition problem}.

\newpage

\bibliographystyle{plain}
\bibliography{ref}

\begin{thebibliography}{10}

\bibitem{Alzantot2019}
M.~Alzantot, Y.~Sharma, S.~Chakraborty, H.~Zhang, C.~Hsieh, and M.~Srivastava.
\newblock {GenAttack}: practical black-box attacks with gradient-free
  optimization.
\newblock In {\em Proceedings of the Genetic and Evolutionary Computation
  Conference}, pages 1111--1119, 2019.

\bibitem{Andriushchenko2020}
M.~Andriushchenko, F.~Croce, N.~Flammarion, and M.~Hein.
\newblock Square attack: A query-efficient black-box adversarial attack via
  random search.
\newblock In {\em Computer Vision}, pages 484--501. 2020.

\bibitem{Bose2019}
A.~Bose.
\newblock {Handwritten Digit Recognition Using PyTorch: Intro to Neural
  Networks}.
\newblock
  \url{https://towardsdatascience.com/handwritten-digit-mnist-pytorch-977b5338e627},
  2019.
\newblock [2024-09-08].

\bibitem{Brunner2019}
T.~Brunner, F.~Diehl, Michael~T. Le, and A.~Knoll.
\newblock Guessing smart: Biased sampling for efficient black-box adversarial
  attacks.
\newblock In {\em IEEE/CVF International Conference on Computer Vision (ICCV)},
  2019.

\bibitem{Chakraborty2021}
A.~Chakraborty, M.~Alam, V.~Dey, A.~Chattopadhyay, and D.~Mukhopadhyay.
\newblock A survey on adversarial attacks and defences.
\newblock {\em CAAI Transactions on Intelligence Technology}, 6(1):25--45,
  2021.

\bibitem{Chang2022}
H.~Chang, Y.~Rong, T.~Xu, W.~Huang, H.~Zhang, P.~Cui, X.~Wang, W.~Zhu, and
  J.~Huang.
\newblock Adversarial attack framework on graph embedding models with limited
  knowledge.
\newblock {\em IEEE Transactions on Knowledge and Data Engineering}, pages
  4499--4513, 2022.

\bibitem{Chen2021}
S.~Chen, C.~Li, and H.~Lin.
\newblock {A Unified View of {cGANs} with and without classifiers}.
\newblock In {\em {Advances in Neural Information Processing Systems:
  Proceedings of the 2021 Conference}}, pages 27566--27579, 2021.

\bibitem{Clare2023}
L.~Clare and J.~Correia.
\newblock {Generating Adversarial Examples through Latent Space Exploration of
  Generative Adversarial Networks}.
\newblock In {\em {Proceedings of the Companion Conference on Genetic and
  Evolutionary Computation}}, pages 1760--1767, 2023.

\bibitem{Darlow2018}
L.~Darlow, E.~Crowley, A.~Antoniou, and A.~Storkey.
\newblock {CINIC-10 Is Not ImageNet or CIFAR-10}, 2018.
\newblock [20-11-2024].

\bibitem{Fan2024}
M.~Fan, Y.~Liu, C.~Chen, and X.~Liu.
\newblock Semiadv: Query-efficient black-box adversarial attack with unlabeled
  images, 2024.

\bibitem{Foster2019}
D.~Foster.
\newblock {\em Generative Deep Learning}.
\newblock O`Reilly Media, Inc., 2019.

\bibitem{explaining_adversarial_examples}
I.~Goodfellow, J.~Shlens, and C.~Szegedy.
\newblock {Explaining and Harnessing Adversarial Examples}.
\newblock In {\em {3rd International Conference on Learning Representations}},
  pages 1--9, 2015.

\bibitem{Gui2023}
J.~Gui, Z.~Sun, Y.~Wen, D.~Tao, and J.~Ye.
\newblock A review on generative adversarial networks: Algorithms, theory, and
  applications.
\newblock {\em IEEE Transactions on Knowledge and Data Engineering},
  35(4):3313--3332, 2023.

\bibitem{Iturriaga2014}
S.~Iturriaga, S.~Nesmachnow, F.~Luna, and E.~Alba.
\newblock A parallel local search in cpu/gpu for scheduling independent tasks
  on large heterogeneous computing systems.
\newblock {\em The Journal of Supercomputing}, 71(2):648–672, October 2014.

\bibitem{Li2012}
D.~Li.
\newblock {The MNIST Database of Handwritten Digit Images for Machine Learning
  Research}.
\newblock {\em IEEE Signal Processing Magazine}, pages 141--142, 2012.

\bibitem{Li2023}
G.~Li, B.~Shi, Z.~Liu, D.~Kong, Yulei Wu, Xiaodan Zhang, Longtao Huang, and
  Honglei Lyu.
\newblock Adversarial text generation by search and learning.
\newblock In {\em Findings of the Association for Computational Linguistics:
  EMNLP 2023}, pages 15722--15738. Association for Computational Linguistics,
  2023.

\bibitem{Liu2022}
S.~Liu, N.~Lu, C.~Chen, and K.~Tang.
\newblock Efficient combinatorial optimization for word-level adversarial
  textual attack.
\newblock {\em IEEE/ACM Transactions on Audio, Speech, and Language
  Processing}, 30:98--111, 2022.

\bibitem{Machin2022}
B.~Machín, S.~Nesmachnow, and J.~Toutouh.
\newblock Multi-target evolutionary latent space search of a generative
  adversarial network for human face generation.
\newblock In {\em Proceedings of the Genetic and Evolutionary Computation
  Conference Companion}, 2022.

\bibitem{Meiner2023}
A.~Mei{\ss}3ner, A.~Fr\"{o}hlich, and M.~Geierhos.
\newblock Keep it simple: Evaluating local search-based latent space editing.
\newblock {\em SN Computer Science}, 4(6):820, 2023.

\bibitem{Mirza2014}
M.~Mirza and S.~Osindero.
\newblock {Conditional Generative Adversarial Nets}, 2014.

\bibitem{Nesmachnow2019}
Sergio Nesmachnow and Santiago Iturriaga.
\newblock {Cluster-UY: Collaborative Scientific High Performance Computing in
  Uruguay}.
\newblock In Springer, editor, {\em Supercomputing}, volume 1151 of {\em
  Communications in Computer and Information Science}, pages 188--202.
  Springer, 2019.

\bibitem{Papernot2017}
N.~Papernot, P.~McDaniel, I.~Goodfellow, S.~Jha, Z.~Celik, and A.~Swami.
\newblock {Practical Black-Box Attacks against Machine Learning}.
\newblock In {\em {Proceedings of the ACM on Conference on Computer and
  Communications Security}}, pages 506--519, 2017.

\bibitem{Singh2020}
R.~Singh, A.~Agarwal, M.~Singh, S.~Nagpal, and M.~Vatsa.
\newblock On the robustness of face recognition algorithms against attacks and
  bias.
\newblock In {\em 34$^{th}$ {AAAI} Conference on Artificial Intelligence},
  pages 13583--13589, 2020.

\bibitem{Szeliski2011}
R.~Szeliski.
\newblock {\em {Computer Vision: Algorithms and Applications}}.
\newblock Springer, 2011.

\bibitem{thys2019fooling}
S.~Thys, W.~{Van Ranst}, and T.~Goedem{\'e}.
\newblock Fooling automated surveillance cameras: Adversarial patches to attack
  person detection.
\newblock In {\em IEEE/CVF Conference on Computer Vision and Pattern
  Recognition Workshops}, pages 49--55. IEEE, 2019.

\bibitem{Vakhshiteh2021}
F.~Vakhshiteh, A.~Nickabadi, and R.~Ramachandra.
\newblock Adversarial attacks against face recognition: A comprehensive study.
\newblock {\em IEEE Access}, 9:92735--92756, 2021.

\bibitem{Volz2018}
Vanessa Volz, Jacob Schrum, Jialin Liu, Simon~M Lucas, Adam Smith, and
  Sebastian Risi.
\newblock Evolving mario levels in the latent space of a deep convolutional
  generative adversarial network.
\newblock In {\em Proceedings of the genetic and evolutionary computation
  conference}, pages 221--228, 2018.

\bibitem{Wang2023}
X.~Wang, Z.~Zhao, C.~Zhang, N.~Bai, and X.~Hu.
\newblock {\em SE-ResNet56: Robust Network Model for Deepfake Detection}, page
  37–52.
\newblock 2023.

\bibitem{Wu2021}
C.~Wu, W.~Luo, N.~Zhou, P.~Xu, and T.~Zhu.
\newblock {Genetic Algorithm with Multiple Fitness Functions for Generating
  Adversarial Examples.}
\newblock In {\em {Congress on Evolutionary Computation}}, pages 1792--1799,
  2021.

\bibitem{Xiao2018}
C.~Xiao, B.~Li, J.~Zhu, W.~He, M.~Liu, and D.~Song.
\newblock Generating adversarial examples with adversarial networks, 2018.

\end{thebibliography}

\end{document}